# SAFE REASONING OVER ONTOLOGIES

Genady Grabarnik
Aaron Kershenbaum
IBM TJ Watson Research center
genadyg@gmail.com, kersh@optonline.net

**ABSTRACT**

As ontologies proliferate and automatic reasoners become more powerful, the problem of protecting sensitive information becomes more serious. In particular, as facts can be inferred from other facts, it becomes increasingly likely that information included in an ontology, while not itself deemed sensitive, may be able to be used to infer other sensitive information.

We first consider the problem of testing an ontology for "safeness" defined as its not being able to be used to derive any sensitive facts using a given collection of inference rules. We then consider the problem of optimizing an ontology based on the criterion of making as much useful information as possible available without revealing any sensitive facts.

## 1. INTRODUCTION

Ontologies are finding increasingly wide use as tools for organizing information. An ontology includes a set of concepts, relationships among them, and properties associated with these concepts. It also includes individuals, relationships among them, and properties they possess, from which we can infer their association with the concepts. As example, we might declare the concepts "person", "parent", and "man". Then, we might define the property "hasChild". We could then define parent as a person which is in a hasChild relationship with another person. It might also include a constraint that a person can be in a hasChild relationship with at most two parents.

The languages used to define ontologies differ from one another in terms of their expressiveness. In general, the more expressive a language is, the more complex the concepts it can express but the more difficult it is to reason in it. By reason we mean testing for consistency, determining the associations of individuals with concepts and determining subsumption relationships among the concepts.

Not all relationships are explicitly stated. Some may be inferred from the other information in the ontology and from the constraints. Thus, for example, if we find that "Hillary hasChild Chelsea", "Bill hasChild Chelsea" and "Senator Clinton hasChild Chelsea" we might infer that at least two of the parents involved in those three statements are in fact the same person. That, together with some other information and inferences, might allow us to infer that Hillary's last name is Clinton and that she is a Senator.

In general, the power of ontologies is that they allow us to infer information from what is explicitly stated. This reduces the amount of information that has to be stated explicitly and also allows us to retrieve information based on concepts in addition to based on specific words or property values. This is in general good, but may also lead to a problem if some of the inferred information is sensitive. For example, while the knowledge base may not explicitly state that some individual is employed by Enron, it may be possible to infer this from other information in the ontology together with the definition of the concept "Enron employee" (i.e., a definition of the criteria for someone being an Enron employee) and the constraints on the relation "isEmployedBy".

Ontology security and privacy are central issues considered by W3C (see for example [D.Wietzner, 2005]). In this paper we introduce notion of safe reasoning over ontologies. We explore how to test if an ontology explicitly contains enough information to infer any facts which are identified as sensitive. We then go on to describe an algorithm for making an ontology safe while still revealing as much information as possible. Our algorithm is based on the application of classical matroid theory (see [E. Lawler, 1976] and [M. Gondran et. al.,1984] for details on matroid theory). From the latest results on matroid theory we point out survey [E. Boros et. all, 2003].

Paper proceeds as follows. We first give a description of the representation of an ontology as a set of triple (RDF presentation). Then, we consider reasoning over ontologies more closely. Section 3 introduces testing an ontology for safeness. Section 4 contains a more formal description of the algorithm. The conclusion section outlines the paper's contributions and areas for future work.

## 2. SAFE REASONING OVER ONTOLOGIES

### 2.1 Presentation of Ontology as RDF triples

We define an ontology, O, as a tuple { I , R , M } , where I, R and M are, respectively, finite sets of individuals, relationships and metadata. In general, M may include characteristics of relations (e.g., symmetry, transitivity) and constraints on relationships (e.g. restrictions on the number of relationships of a given type that may exist between individuals). While standard definitions of ontologies often make a distinction between individuals and classes, for simplicity we will not do so here as this distinction does not add anything to the discussion. Thus, the relationship
    ( individual  isMemberOf  class )
is treated no differently than any other relationship and metadata rules may apply to relationships among both individuals and classes without distinction.

A relationship, r, in R is expressed as a set of triples of the form,
    ( subject , property , object )

where 'subject' is an individual, 'property' is a specific type of relationship, and 'object' is an expression composed of individuals and the logical operators AND, OR and NOT.

Examples of relationships are:
   ( Jim isMemberOf man )
   ( man isEquivalentTo (person AND male ) )
   ( American isSubsetOf person )
Each of these triples can be thought of as a fact about the individuals involved.

Pieces of metadata, m in M, are expressed as triples of the form,
   ( property, constraint , value )
where property is the middle member of a relationship triple, value may be a property or constant, and constraint is a member of
{ < = > inverseOf subPropertyOf disjointFrom is }.

Examples of metadata are:
   ( isSubsetOf is transitive ); i.e., the property isSubsetOf is transitive
   ( name = 1 ) ; i.e., every individual must have exactly one name
   ( spouse < 2 ) ; i.e., everyone must have at most one spouse
   ( parentOf inverseOf childOf )
   Types of metadata give rise to inference rules, for example,
      ( ancestorOf is transitive )
   allows us to infer that if
      ( Adam ancestorOf Bob ) and
      ( Bob ancestorOf Carl )
   then
      ( Adam ancestorOf Carl )
   Another example is,
      ( person is (man OR woman) )
      ( man disjointFrom woman )
      ( Adam is person )
      ( Adam is ( NOT woman ) )
   allows us to infer
      ( Adam is man )
   We extend the definition of an ontology to include restricted relations of the form
FOR_ALL individuals, i, in class c, there exists an individual, j, in class D such that
         ( i property j )
and
FOR_ALL individuals, i, in class C, if there exists an individual, j, such that
         ( i property j )
then j is a member of class D.

### 2.2 Computing the closure of a set of relations

The first problem we consider is how to compute the closure of a set of relations; i.e., the total set of relations (facts), F(R), that can be inferred from a given set of relations, R, and the inference rules implied by metadata M.

If M is simple enough, this problem is simple too. Suppose that M only contains,
- ( isSubsetOf is transitive )
- ( isEquivalentTo is transitive )
- ( isEquivalentTo is symmetric )

Then, given a set of relations of the form
- ( x isSubsetOf y )
- ( w isEquivalentTo z )
- ( i isA C )

F(R), the closure of a set of relations, R, can be computed by considering a graph, G, with edge set R; i.e., the triples in R define the set of edges of G and the endpoints of these edges define the set of nodes.

In this case, the only inferences we can make are membership inferences; i.e., we can infer whether some set is equivalent to or a subset of another set and we can infer whether an individual is a member of a set. Note that this formulation allows us to distinguish between individuals and sets. This can be done in a number of ways including allowing the ontology to include relations of the form
- ( C isA Set )

,where Set is a distinguished "individual".

The problem of determining the closure of a set of relations, R, in this case is then just becomes one of identifying the reachability set of each node, n, in G; i.e., determining for which set of nodes, s, a path exists from n to s. This is easily computed by using breadth first search.

In a more general case, we may have other transitive relations, for example, isPartOf, e.g.,
- (USA isPartOf NorthAmerica)
- (StatePennsylvania isPartOf USA)
- (CityPhiladelphia isPartOf StatePennsylvania )

The problem of determining membership in this case can still be solved using a simple search algorithm, but it now must be sensitive to the fact that paths must be comprised of properties which are of the same type. Note that this can be extended to the case where some different types of properties can interact to form paths by declaring all such groups of properties as subProperties of a single transitive property.

The most general case, where more complex inferences are possible, is discussed in the following section.

### 2.3 Testing an Ontology for Safeness

The basic problem of testing a part of ontology for safeness is closely related to that of determining its closure. Specifically, we are given
- An ontology, O = { I , R , M }
- A subset, $R_s$, of R which contains all sensitive relations (facts)
- A subset, Q, of R which is to be tested for safety

We say that Q is safe if its closure, F(Q), does not contain any fact in $R_s$.

Thus, if we can efficiently determine F(Q), we can answer the question of whether Q is safe. It is clear that this can be done in the simple cases discussed in the previous section. We now consider a more general case. In the preceding section we assumed that the only inference mechanism was transitivity. Thus, we inferred facts from other individual facts. This leads us to simply look for paths comprised of relationships between specific pairs of individuals. Now suppose that facts are inferred from groups of other facts. For example, suppose we are given the triples

    $T_1$: ( A isEquivalentTo  (B AND C) )
    $T_2$: ( A  is subSetOf D )
    $T_3$: ( E isEquivalentTo (B AND (C AND D)))

The (sensitive) fact that

    $T_4$: ( A is subSetOf E )

can be inferred from $T_1$, $T_2$ and $T_3$, but not from any two of them.
We would thus say the sub-ontologies ( $T_1$ , $T_2$ ) , ($T_1$ , $T_3$) and ($T_2$ , $T_3$) , are safe but that the sub-ontology ( $T_1$ , $T_2$ , $T_3$ ) is not safe.

In general, if we are given for each $r_{si}$ in $R_s$, one or more sets of relations $M_{si}^k$, such that $r_{si}$ can be inferred from $M_{si}^k$, but cannot be inferred from any subset of $M_{si}^k$, then we say that any sub-ontology containing all the relations (facts) in $M_{si}^k$ is not safe. If we now consider all $M_{si}^k$, we can define a safe sub-ontology as any set of relations which does not contain all the members of any $M_{si}^k$.

In the general case, finding all such $M_{si}^k$ is impractical by any currently known means, although it is theoretically possible for many existing logic systems, (e.g., OWL-DL) by expanding a sufficient number of tableaux. In the case of an ontology defined by Horn clauses, however, the $M_{si}^k$ are given explicitly or can be derived in a reasonable amount of time using the procedures described in  [see Baader, et. el., 2003 for details ]. In the following section, we will assume that we have been given the $M_{si}^k$, which is reasonable at least for cases where the relations are described by Horn clauses or where the ontology is otherwise simple enough to be analyzed.

### 2.4 Finding an  Optimal Safe Ontology

We now turn to the problem of finding the "best" safe ontology. The simplest notion of best is that we retain as many relations as possible without revealing any sensitive information. The techniques we describe can be extended to the case where the relations have weights and we seek a safe ontology of maximum weight. We will discuss both of these problems, but for clarity usually discuss the first.

We are thus given an ontology, O, and a set of sensitive relations $R_s$ = { $r_{si}$ }. For each $r_{si}$ , we are given $M_{si}^k$ , the minimal sets of relations required to infer $r_{si}$ . (Note that it is possible for there to be more than one $M_{si}^k$ for a given $r_{si}$ ). We wish to find a maximum cardinality set of relations, $R^*$ such that $R^*$ does not include all the relations in any of the $M_{si}^k$. For simplicity in the discussion that follows, we refer to the sets $M_{si}^k$ simply as $M_j$. since their relationship to the specific $r_{si}$ is not relevant to our approach.

The approach we take is to find a maximum cardinality intersection of matroids [F. Maffioli, 1975], [E. Lawler 1973]. A matroid ***M***( E , ***F*** ) is defined by a (here finite) set of elements E and a family ***F*** of independent subsets, F, of E where the independent sets have the properties that
1. Every subset of an independent set is also independent
2. If there are two independent sets, $F_k$ and $F_{k+1}$, of cardinalities k and k+1, respectively, then there exists an element $e_i$ which is a member of $F_{k+1}$ but is not a member of $F_k$ and such that $F_k \cup e_i$ is an independent set.

This second property leads to very simple algorithms for finding maximum cardinality independent sets in matroids. One need only find elements which are independent of those already selected, with the assurance that no element selected will prevent us from finding an independent set of higher cardinality if such a set exists. Indeed, a much stronger result exists [M. Gondran et. al.,1984]. If there are weights associated with the elements and we consider elements in order of weight largest first (Greedy Algorithm), we are guaranteed to find an independent set of maximum weight. This result is the key to extending our approach to the case where relations have weights.

It is easy to see that each pair ( $M_j$ , $F_j$ ) , where $F_j$ includes all proper subsets of $M_j$, forms a matroid since every subset of an independent set is independent and any larger independent set, $F_{k+1}$ , must contain an element not contained in a smaller set , $F_k$, and $F_k$ must be missing at least two elements of $M_j$, so adding a single element to $F_k$ could not complete $M_j$ and hence would still leave $F_k$ independent.

Thus, we can associate a matroid ***M***$_j$ with each $M_j$. It is trivial to find a maximum cardinality (or maximum weight) subset of $M_j$. But we want to find a single set of relations which are, simultaneously, independent in all the $M_j$. This is called an independent set in an intersection of matroids.

Formally, given k matroids ***M***$_1$ , ***M***$_2$ , … ***M***$_k$ all defined over the same element set, E, we define … ***M***$_I$, the intersection of these matroids, as ***M***$_I$ = ( E , ***F***$_I$ ) , where a subset, F, of E is a member of ***F***$_I$ if and only if it is independent in all the individual matroids.

A polynomial bounded algorithm [E. Lawler, 1973] exists to find an independent set of maximum cardinality in the intersection of two matroids. (The algorithm can be extended to find intersections of maximum weight). It relies on the concept of an alternating chain and is an extension of the algorithm for finding maximum cardinality independent sets in a single matroid.

The algorithm begins by selecting elements one at a time, maintaining independence in both matroids, until no further elements can be selected. Unlike the case with a single matroid, however, it is no longer true that one can guarantee finding a maximum cardinality intersection in this way. Consider, for example two matroids, ***M***$_1$ and ***M***$_2$ , both defined on the set of elements
  E = { $e_1$ , $e_2$ , $e_3$ , $e_4$ , $e_5$ }

We define $F_1$ as
$$\{ (e_1, e_2, e_4), (e_1, e_3, e_4), (e_1, e_2, e_5), (e_1, e_3, e_5) \}$$
and all their subsets, and we define $F_2$ as
$$\{ (e_1, e_2, e_5), (e_1, e_4, e_5), (e_2, e_3, e_5), (e_2, e_4, e_5), (e_3, e_4, e_5) \}$$
and all their subsets.

Suppose we begin by selecting $(e_2, e_4)$, which is independent in both matroids. This set is maximal since we cannot add any elements to it, but it is not of maximum cardinality since $(e_1, e_2, e_5)$ is independent in both matroids. What is required is to remove $e_4$ from $(e_2, e_4)$ and then add $e_1$ and $e_5$. Specifically, we can add $e_1$ maintaining independence in $M_1$, but destroying independence in $M_2$. Then we remove $e_4$, restoring independence in $M_2$. Then, we add $e_5$ maintaining independence in $M_1$, and since it also happens to maintain independence in $M_2$, we do not need to remove anything more. We have found a larger independent. When we try to do this again, starting with the new independent set $(e_1, e_2, e_5)$, we find we cannot do this again.

We call the sequence of modifications "add $e_1$, remove $e_4$, add $e_5$" an augmenting path as it augments an independent set, creating a larger one. It is clear that if an augmenting path exists with respect to an independent set F, then F is not of maximal cardinality. It has also been proven [E. Lawler, 1973] that if an independent set F is not of maximal cardinality then an augmenting path must exist with respect to F.

The algorithm for finding an augmenting path with respect to some F (if one exists) is essentially a traversal. We begin by finding some element $e_j$ which is independent of F in $M_1$. If no such $e_j$ exists, then no augmenting path exists. If such an $e_j$ exists and it is also independent of F in $M_2$, then $e_j$ is itself an augmenting path. If $e_j$ is not independent of F in $M_2$, then $e_j$ forms a circuit (minimally dependent set) with F and independence in $M_2$ can be restored by removing any other element of this circuit. Having done so, it may now be possible to find some element which can be added to F while still maintaining independence in $M_1$. We continue along the same lines, alternately adding and deleting elements, each time maintaining independence in $M_1$ when adding to F and restoring independence in $M_2$ by deleting an element until we either find that an added element maintains independence in $M_2$ as well as $M_2$ or that no such element can be found. It has been shown [E. Lawler, 1973] that this process terminates in polynomial time because elements in F which are deleted, never reenter F during any given augmentation. Indeed, if a breadth first search is used in the traversal, no node or edge is visited more than once.

Unfortunately, we will usually be intersecting far more than two matroids and the augmenting path algorithm, when extended to more than two matroids is not guaranteed to converge in polynomial time. We can, however, adapt it so that it will perform efficiently in many realistic cases.

The first step in constructing an efficient algorithm is to transform the problem from one of intersecting k (usually a large number) matroids to one of intersecting three matroids.

We first make j copies of each element in E, one for each $M_j$. We can then easily find independent sets in each of the matroids separately, but after doing so we must enforce the restriction that if we used copy j of $e_i$ in the independent set from $M_j$, we must also use $e_i$ in the independent sets for all other each $M$'s as well. Thus, we have transformed the k-intersection problem in a matroid $M$ with m elements into one of finding a maximum cardinality (or maximum weight) independent set in a matroid $M^*$ with km elements but with an additional condition, called a parity condition, that all copies of a given element be included in any solution.

We now remove the parity condition by defining two additional matroids on the elements of $M^*$. First, corresponding to each element $e_{ij}$ in $M^*$ we define a new element, $a_{ij}$. We now define a new matroid,

$M_1^{**} = (E^{**}\ F_1^{**})$, where $E^{**} = \{e_{ij}\} \cup \{a_{ij}\}$ and F is in $F_1^{**}$ if all e in $F \wedge E_j$ ( the $j^{th}$ copies of E ) are independent in $M_j$. Thus, $M_1$ enforces the constraints in the original matroids. To enforce the parity constraint we define

$M_2^{**} = (E^{**}\ F_2^{**})$, where F is in $F^{**}_2$ if for all i and j, F does not include both $e_{ij}$ and $a_{ij}$.

$M_3^{**} = (E^{**}\ F_3^{**})$, where F is in $F^{**}_3$ if for all i and j ( j = 1, 2, … k), F does not include both $e_{ij}$ and $a_{i,j+1}$ for j < k and does not contain both $e_{ik}$ and $a_{i,1}$.

The effect of the constraints in $F_2^{**}$ and $F_3^{**}$ is to allow a full set of $e_{ij}$'s for a given i or a full set of $a_{ij}$'s for that given i, but not both. If the $e_{ij}$'s have weights $w_{ij}$ associated with them, we can then seek a maximum weight independent set in the intersection of these three matroids. If we add a large enough constant, C, to the weights of all the $a_{ij}$'s and $a_{ij}$'s, then the maximum weight intersection will also be maximum cardinality. Thus, the optimal solution will respect the parity condition, since it will contain full sets of $e_{ij}$'s.

Now, we "only" have to solve the problem of finding a maximum weight intersection over the intersection of three matroids. This problem is known to be NP-complete [R.M. Karp, 1972] but is nevertheless tractable in many cases of interest. We have the advantage here that the three matroids involved are partition matroids; i.e., matroids where the independence condition is simply that the number of elements in each set partitioning E is limited. This makes finding independent sets very easy.

To extend the augmenting path algorithm from the problem of finding 2-matroid intersections to the problem of finding 3-matroid intersections we can still begin an augmenting path with an element $e_j$ which is independent of F in $M_1$. If $e_j$ which is also independent of F in $M_2$ and $M_3$, then $e_j$ is an augmenting path. If $e_j$ is not independent of F in $M_2$ but is independent of F in $M_3$, then we can, as before, restore independence in $M_2$ by removing any element in the circuit formed by $e_j$ in $M_2$. We can handle the case where $e_j$ is not independent of F in $M_3$ but is independent of F in $M_2$ in an analogous way.

The new situation we must handle is where $e_j$ is not independent of F in $M_2$ or in $M_3$. In this case there are two possibilities we must consider. We have just formed two cycles, one in $M_2$ and one in $M_3$. It is possible that these two cycles share an element $e_k$ ($\gamma e_j$). In this case we could remove $e_k$ and restore independence in both $M_2$ and $M_3$. But it is also possible to restore independence in $M_2$ and $M_3$ by removing two separate elements $e_{k2}$ and $e_{k3}$. Indeed, if no $e_k$ as above exists, this would be necessary. It is always possible to find such $e_{k2}$ and $e_{k3}$, but by removing two elements and adding only one, we have reduced the size of the independent set. We must thus consider adding another element, $e_{j2}$ which is independent of the other selected elements in $M_1$; i.e., $E_p - e_j + e_{k2} + e_{k3}$. It is possible that this will result in a loss of independence in $M_2$ and $M_3$ or even both. In this case, we must remove additional elements to restore independence.

Because we are now considering exchanging groups of elements, rather than just single elements, it is now possible that elements may participate in more than one path. Because of this, the number of paths we may need to consider can, in the worst case, become exponentially large. In practice, however, this is not likely to happen because the exchange of elements will make the solution worse. The only reason we are even considering such exchanges is that they may lead to a better solution which was not available before the exchange. The more we exchange, the less likely this becomes. Note that once we find an augmenting path the only reason we would continue to search for alternatives is that they might lead to a solution better than the one already found. If we can bound the value of the best solution reachable from a given partial augmentation, we can eliminate it from consideration if the bound is already worse than the augmentation we have already found.

Thus, the only case that is problematical is when we cannot find any augmentation at all. We know that we will eventually encounter this because eventually we will have found a maximum cardinality intersection. In this case, however, we may still find that the overall effort is acceptable. In order to be sure, however, we need to implement this algorithm and run experiments with it.

We now give a more formal description of our algorithm, along the general lines of the algorithm presented in [M. Gondran et. al.,1984] for intersecting two matroids. We first define a graph, B, known as the border graph associated with an independent set with p elements, $I_p$, with the property that an augmenting sequence that takes us from $I_p$ to $I_{p+1}$ corresponds to one or more paths in B. B is a bipartite graph whose node set is E, the element set for the three matroids. The nodes of B are partitioned into the sets $I_p$ and $E-I_p$. For $e_i \chi I_p$ and $e_j \chi E-I_p$, there is a directed edge ($e_j$, $e_i$) in B if $e_j$, when added to $I_p$, forms a cycle $C_j^{(1)}$ in $M_1$ and if $e_i \chi C_j^{(1)}$. Similarly, there is a directed edge ($e_i$, $e_j$) in B if $e_i$, when added to $I_p$, forms a cycle $C_j^{(2)}$ in $M_2$ and if $e_j \chi C_i^{(2)}$ or if $e_i$, when added to $I_p$, forms a cycle $C_j^{(3)}$ in $M_3$ and if $e_j \chi C_i^{(3)}$. We refer to edges of B which are based on a cycle in $M_1$ as type-1 edges and in general to edges based on cycles in $M_k$ as type-k edges.

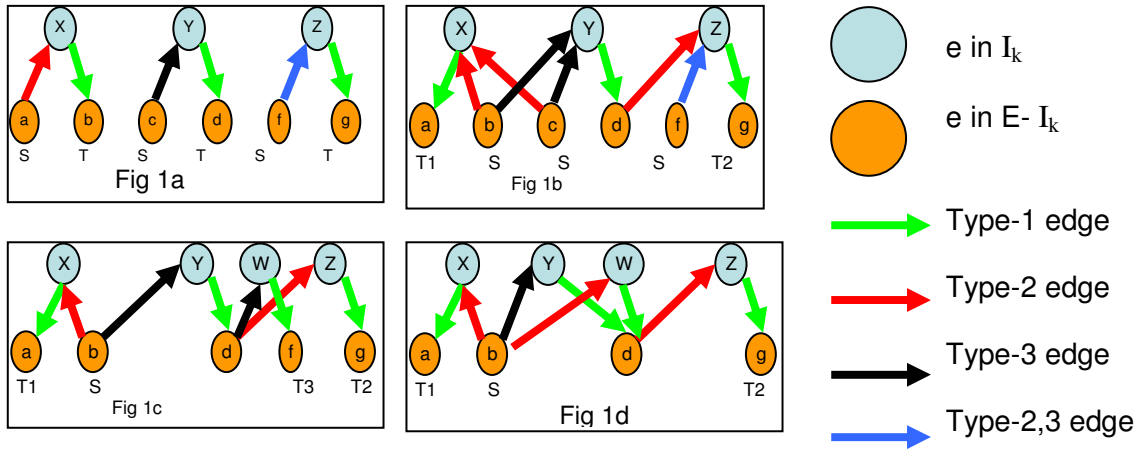

Fig 1 Illustrates the algorithm for finding an augmenting tree. Fig 1a illustrates the process of creating boundary graph B. Fig. 1b illustrates building paths to resolve dependency for one of the augmenting tree branches. Fig 1c.illustrates branching in the augmenting tree when adding b creates 2 cycles in $M_2$ and $M_3$. Fig 1d illustrates resolution of the "cycle" created by b,Y,W, and d.

An augmenting sequence corresponds to one or more paths in B starting at some node, $e_1 \chi E-I_p$ ,with no incoming edges and ending at one or more nodes in E-$I_p$ with no outgoing edges. In the simplest case, $e_1$ has neither incoming nor outgoing edges. In this case, it forms no cycles with $I_p$ in any of the three matroids and $e_1$ is an augmenting path by itself; i.e. it can be added to $I_p$ to form $I_{p+1}$. Such nodes are isolalated nodes in B.

The next most simple case is where $e_1$ has no incoming edges (i.e., it does not form a cycle in $M_1$ added to $I_p$) but does form a cycle in $M_2$. In this case, if we add $e_1$ to $I_p$ we must remove some $e_j$, in the cycle it forms in $M_2$. These are precisely the $e_j$, connected to $e_1$ via a type-2 edge B. We must then find an edge from $e_j$ to some node $e_k$ in $I_p$ where $e_j$ is part of the cycle formed by $e_k$ in $M_1$. In Figure 1b, the path from b to X to a is such a path. An analogous statement can be made when $e_1$ forms a cycle in $M_3$; the path from c to Y to d is an example. It is also possible that $e_1$ has no incoming edges but forms a cycle in $M_2$ and one in $M_3$. If there is a single node, $e_j$, which is in both of these cycles, then we can treat this case like the previous one; i.e., we can add $e_1$, remove $e_j$, and add some node $e_k$ to which includes $e_j$ in the cycle $e_k$ forms with $I_p$ in $M_1$. The path from f to Z to g is an example of such a path.

Augmenting paths may contain more than 3 nodes. The path c-Y-d-Z-g in Figure 1b is an example of such an augmenting path. In this case, removing Y breaks the cycle in $M_3$ formed by adding c, but when we add d it forms a new cycle in cycle in $M_2$. Removing Z breaks this cycle and adding g forms no new cycles, completing the augmentation.

It is also possible that adding $e_1$ forms a cycle in $M_2$ and one in $M_3$. but that there is no single $e_j$ like g in Figure 1a that is part of both cycles. An example of this is b in Figure 1b. In this case we need to resolve the two cycles separately. The cycle in $M_2$ is resolved by the path b-X-a and the cycle in $M_3$ is resolved by the path b-Y-d-Z-g. So in this case the augmenting "path" is actually 2 paths. We refer to the union of such "paths" as augmenting trees.

This problem can get worse. In Figure 1c, we find that the cycle that b forms in $M_2$ is again resolved by the path b-X-a, but the cycle b forms in $M_3$ again starts with b-Y-d, but now we must resolve two separate cycles formed by d. In this case, the problem is resolved by the simple paths d-W-f and d-Z-g and these four paths form a tree. While this is somewhat more complicated than the preceding cases, it is still straightforward algorithmically because we are just scanning outward from b and forming a tree. The effort is still polynomial.

If these were the only cases we needed to consider, we would be very happy indeed because we would have just given a polynomial bounded algorithm to solve an NP-complete problem, thereby proving P=NP. Unfortunately, there is one more case to consider. In Figure 1d, b forms two cycles. The cycle in $M_2$ can be resolved either through X or through Y. The cycle in $M_3$ must be resolved through W. If we try to resolve the cycle in $M_2$ through Y, we must continue to d. But the cycle in $M_3$ must be resolved through W and the only way out of W is also through d. The paths b-Y-d and b-W-d are not compatible. If we use them both, we will not succeed in creating an augmentation. In particular, we will not increase the size of the independent set. So we must use the d-X-a path to resolve the cycle in $M_2$. This leaves d free to participate in the b-W-d path to resolve the cycle in $M_3$. A similar problem arises when we try to resolve the cycle in $M_2$ at d. If we try to resolve it through X, we will find that we must go through a again and we will fail to produce a valid augmentation. We thus must use the d-Z-f path to resolve this cycle. In this particular case, because the problem was small we are able to see how to resolve the problem. In the general case, however, we may have to wait a long time to do so. In general we have to hold both possibilities open until the problem is completely resolved. This leads us to the following algorithm which is in the worst case exponential.

Step 0: Start with an empty intersection, $I_0$. Set k=0.

Step 1: Form the border graph, B, based on the current intersection $I_k$.

Step 2: Find an augmenting tree, $T_k$ in B. Since we are seeking the maximum weight intersection, we seek an augmenting tree of maximum weight. In order for this to be a tree, rooted at some starting element $e_1$ there must be at most one path leading into each node. It is thus in general necessary to allow the $e_j$ to have multiple labels. We define a label as a tuple ( S , W ), where S is the set of $e_j$ in the path from $e_1$ and W is the total weight of all $e_j$ in the path. We say a label ($S_1$, $W_1$) dominates another label ($S_2$, $W_2$) if $S_1$ is a subset of $S_2$) and $W_1$ [ $W_2$). We find an augmenting tree rooted at $e_1$ by labeling $e_j$ from previously labeled $e_k$. All paths in the tree must terminate in $e_k$ with degree 0. This resolves all cycles formed while doing the augmentation. If no such path can be found, the current $I_k$ is of maximum cardinality. If no path with positive weight can be found, the current $I_k$ is of maximum weight. In either of these cases, we are done. Otherwise, proceed to Step 3.

Step 3: Augment $I_k$ using $T_k$ i.e., add to $I_k$ all $e_j \varpi I_k$ and remove from $I_k$ all $e_j \chi I_k$. Return to Step 1.

## 3. CONCLUSION

We have presented a formulation of the problem of maintaining safety in an ontology and an efficient algorithm for finding ontologies which maximize the amount of information included without compromising any sensitive facts via inference. We have shown that the problem can be solved within the context of matroid theory by finding a maximum weight intersection of three matroids. While this problem is, in general NP-complete, we believe our algorithm can produce near optimal solutions in a reasonable amount of time.

Our plan is to first implement this algorithm and test its computational complexity on realistic problems. We will then consider extending the formulation to include more expressive ontologies and to provide stronger guarantees on the quality of the solution obtained. We also plan to experiment further, using additional ontologies which exhibit a wider variety of structures in order to explore the types of features which make our approach most attractive.